\documentclass[letterpaper, 10 pt, conference]{ieeeconf}  

\IEEEoverridecommandlockouts                              

\usepackage[sorting=none, style=numeric-comp, maxbibnames=99]{biblatex}

\usepackage{color}
\usepackage{epsfig}
\usepackage{graphicx}
\usepackage{algorithm,algorithmic}

\usepackage{adjustbox}
\usepackage{array}
\usepackage{booktabs}
\usepackage{colortbl}
\usepackage{float,wrapfig}
\usepackage{framed}
\usepackage{hhline}
\usepackage{multirow}
\usepackage{subcaption} 
\usepackage[font=small]{caption}
\usepackage[percent]{overpic}

\usepackage{amsmath,amsfonts,amssymb}
\usepackage{amsthm} 
\usepackage{bm}
\usepackage{nicefrac}
\usepackage{microtype}
\usepackage{contour}
\usepackage{courier}

\usepackage{changepage}
\usepackage{extramarks}
\usepackage{fancyhdr}
\usepackage{lastpage}
\usepackage{setspace}
\usepackage{soul}
\usepackage{xspace}
\usepackage{cuted}
\usepackage{fancybox}
\usepackage{afterpage}

\usepackage[breaklinks=true,colorlinks,backref=True]{hyperref}
\hypersetup{colorlinks,linkcolor={black},citecolor={MSBlue},urlcolor={magenta}}
\usepackage{url}
\usepackage{quoting}
\usepackage{epigraph}

\usepackage{enumerate}
\usepackage{paralist,tabularx}
\usepackage{comment}
\usepackage{pdfpages}
\usepackage{caption}
\usepackage{subcaption}

\usepackage{pifont}

\usepackage{MnSymbol}

\usepackage{enumitem}

\makeatletter
\DeclareRobustCommand\onedot{\futurelet\@let@token\@onedot}
\def\@onedot{\ifx\@let@token.\else.\null\fi\xspace}

\def\etal{et al\onedot}

\makeatother

\definecolor{MyDarkBlue}{rgb}{0,0.08,1}
\definecolor{MyDarkGreen}{rgb}{0.02,0.6,0.02}
\definecolor{MyDarkRed}{rgb}{0.8,0.02,0.02}
\definecolor{MyDarkOrange}{rgb}{0.40,0.2,0.02}
\definecolor{MyPurple}{RGB}{111,0,255}
\definecolor{MyRed}{rgb}{1.0,0.0,0.0}
\definecolor{MyGold}{rgb}{0.75,0.6,0.12}
\definecolor{MyDarkgray}{rgb}{0.66, 0.66, 0.66}
\definecolor{MyPink}{rgb}{1, 0.75, 0.79}
\definecolor{GreenStarColor}{rgb}{0.54, 0.84, 0.41}
\definecolor{MSBlue}{rgb}{0, 0.35, 0.49}

\def\OURS{HULA\xspace}

\DeclareMathOperator*{\argmax}{argmax} 

\title{\LARGE \bf Decision Making for Human-in-the-loop Robotic Agents via \\ Uncertainty-Aware Reinforcement Learning}

\author{
Siddharth Singi\authorrefmark{1}\authorrefmark{2}, %
Zhanpeng He\authorrefmark{1}\authorrefmark{3}, %
Alvin Pan\authorrefmark{3}, %
Sandip Patel\authorrefmark{2}, %
Gunnar A. Sigurdsson\authorrefmark{4},\\%
Robinson Piramuthu\authorrefmark{4}, %
Shuran Song\authorrefmark{3} and %
Matei Ciocarlie\authorrefmark{2}\\ %
\url{https://roamlab.github.io/hula/}
\thanks{\authorrefmark{1} denotes joint first authorship.}
\thanks{\authorrefmark{2}Dept. of Mechanical Engineering, \authorrefmark{3}Dept. of Computer Science, Columbia University, New York, USA}
\thanks{\authorrefmark{4}Amazon Alexa AI}
\thanks{This work was supported in part by an award from the Columbia Center of Artificial Intelligence Technology.}
}
 \AtBeginBibliography{\small}

\begin{document}

\maketitle
\thispagestyle{empty}
\pagestyle{empty}

\begin{abstract} In a Human-in-the-Loop paradigm, a robotic agent is able to act mostly autonomously in solving a task, but can request help from an external expert when needed. However, knowing when to request such assistance is critical: too few requests can lead to the robot making mistakes, but too many requests can overload the expert. In this paper, we present a Reinforcement Learning based approach to this problem,  where a semi-autonomous agent asks for external assistance when it has low confidence in the eventual success of the task. The confidence level is computed by estimating the variance of the return from the current state. We show that this estimate can be iteratively improved during training using a Bellman-like recursion. On discrete navigation problems with both fully- and partially-observable state information, we show that our method makes effective use of a limited budget of expert calls at run-time, despite having no access to the expert at training time. 
\end{abstract}

\section{Introduction}
\label{sec:intro}

Deep Reinforcement Learning (DRL) has shown great progress in learning decision-making for complex robotic skills \cite{kroemer2019manipulation, Daniel2016reps, peter2008rlmotorskills, theodorou2010reinforcementklo} using experiences collected by a robotic agent exploring an environment and receiving reward signals. Traditional RL agents  use a policy learned during training in order to act autonomously at deployment. However, even a well-trained agent can encounter situations when deployed that are hard to make decisions for, for reasons such as partial state observability, uncertain dynamics, changes in state distributions between training and testing, etc.

The Human-in-the-Loop (HitL) paradigm has been developed in robotics precisely for situations where an agent can act autonomously most of the time, but would still benefit from receiving assistance from an available (tele-)operator, usually assumed to be a human expert. This paradigm is particularly powerful if the agent itself makes the decision of when to request assistance, thus freeing the operator from having to monitor task progress. However, this approach gives rise to a critical decision-making problem on the agent's part: when to request help? Too few
such requests can lead to the robot making mistakes, but too many
requests will overload the expert and lose the benefit of semi-autonomous operation.

\begin{figure}
    \centering
    \includegraphics[width=\linewidth]{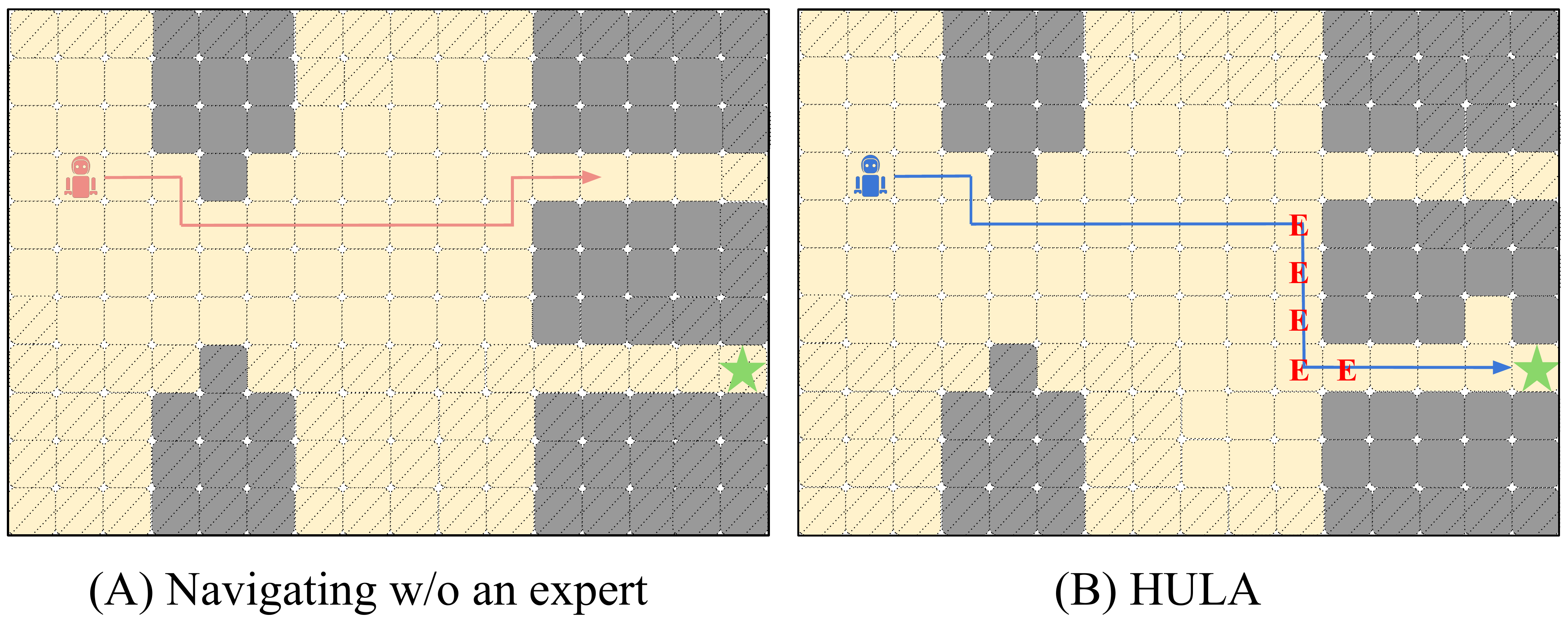}
    \caption{\textbf{An illustration of HULA, the method we propose in this paper.} In a partially-observable environment, an agent without the help of an expert (A) cannot localize itself accurately due to partial observability, goes down the wrong passage and fails to reach the target. A \OURS agent (B) decides to request assistance from an available external expert in the states marked with a red \textcolor{red}{E} and achieves the goal (denoted by a green star \textcolor{GreenStarColor}{$\bigstar$}).  In both cases, the agent can only observe a 5x5 grid around its current location; shadow areas represent unobserved regions throughout the navigation.}
    \label{fig:candy}
    \vspace{-1.5em}
\end{figure}

In this paper, we propose a DRL-based method for a HitL agent to make the critical determination of when to request expert assistance. We posit that the best moment to request such assistance is when the agent is highly uncertain in the successful outcome of the task. From an RL perspective, \textit{we relate this uncertainty to the variance of the return from the current state, as perceived by the agent}. Numerous RL algorithms provide methods for the agent to estimate the expected return from a given state. We show that similar methods can be used to estimate the variance of return as well during the training process. At deployment time, the agent can then request expert assistance when its estimate of the return variance from the current state falls below a given threshold. We dub our method HULA, for Human-in-the-loop Uncertainty-aware Learning Agent, and illustrate its operation in Fig.~\ref{fig:candy}.

Critically, HULA does not need to make any calls to the expert during training. This stands in contrast to a standard RL approach, where an agent could learn how to use an expert simply by making numerous such calls at train time. Nevertheless, our method is able to make effective use of a limited budget of expert calls in order to improve task performance. We summarize our key contributions as follows:
\begin{compactitem}
    \item To the best of our knowledge, we are the first to propose a method for an HitL RL agent to learn how to effectively budget its interactions with a human expert at deployment time, without needing any expert calls during training.
    \item We show that the variance of return, which can be estimated during training using Bellman-like equations, is an effective measure for agent uncertainty and can be used at deployment to determine when to request assistance.
    \item Our experiments on discrete navigation problems with both fully- and partially-observable state information show that our method is as effective or better in managing its budget of expert calls compared to a standard learning approach that also makes thousands of expert calls during training.
\end{compactitem}

\section{Related Work}

Our work extends standard reinforcement learning algorithms to human-in-the-loop policy that solves robotics tasks with the help of humans. Researchers have investigated how to learn such a policy. For example, Arakawa \etal propose DQN-TAMER \cite{arakawa2018dqn}, which is a method that incorporates a human observer model by real-time human feedback during training. 
Expected Local Improvement (ELI) \cite{mandel2017add} trains a state selector that suggests states requiring to query new actions from human experts.
PAINT \cite{xie2022ask} learns a classifier to identify irreversible states and query the expert in case entering such states cannot be avoided.
Hug-DRL \cite{wu2021human} leverages a  control transfer mechanism between humans and automation that corrects the agent’s unreasonable actions during training by human intervention, where it has demonstrated significant potential in autonomous driving applications.
While these methods exploit human intervention during training, our work relies on the internal uncertainty of an agent and only uses an expert in test time.

\OURS explicitly estimates the uncertainty of an RL agent and use it to make decisions about requesting an expert's assistance. 
Prior work has explored representing the uncertainty of outcomes during RL training \cite{Jaderberg2016ReinforcementLW, Mohamed2015VariationalIM, Pathak2017CuriosityDrivenEB}.
For instance, Kahn \etal use the variance of the dynamics model to represent the uncertainty of collision to avoid damage from high-speed collisions 
\cite{kahn2017uncertainty}. Ensemble Quantile Networks (EQN) \cite{hoel2021ensemble} estimates both aleatoric and epistemic uncertainties via the combination of quantile networks(IQN) \cite{dabney2018implicit} and randomized prior function(RPF) \cite{osband2018randomized}.
These works use the uncertainty of an agent to perform task-specific decision-making, while our work differs in using it to collaborate with a human.
Moreover, our work represents an agent's uncertainty using the variance of returns, which is a direct indicator of task performance.

The closest work to ours is RCMP, which learns a HitL policy that queries the expert when the epistemic uncertainty estimated by the variance of multiple value functions is high \cite{da2020uncertainty}. However, RCMP requires expert queries during training, whereas our work learns an uncertainty model based on the variance of return and does not query an expert in train time.

\section{Method}
\label{sec:method}

Consider a problem for an autonomous agent formulated as a standard Markov decision process (MDP). A general MDP is defined as a tuple of $(S, A, r, p)$, where $S$ represents the state space, $A$ represents the action space, $r(s, a)$ is a reward function that evaluates immediate reward of action $a \in A$ in a state $s \in S$, and $p$ represent a transition distribution $p(s_{t+1}| s_{t}, a_{t})$. The goal of solving an MDP is to learn a policy $\pi(a|s)$ that maximizes its expected return $\mathbb{E}_{\pi}[R]$, where $R=\sum\gamma^{t}r(s_t, a_t)$ and $\gamma$ is a discount factor.

Going beyond the standard MDP formulation, we now assume the availability of an expert that can give instructions to the robotic agent. When queried from a given state $s_t$, this expert can  directly provide an action $a_{exp}(s_t)$ for the robot to execute. We assume that the expert always provides high-quality advice, i.e. a policy of always following expert-provided actions from all states will achieve a satisfactorily high return on the MDP problem above. However, such a policy would be impractical: we assume that the expert has limited bandwidth or availability, thus it is desirable for the agent to balance the goal of achieving high returns with the goal of not overloading the expert with requests. Such a scenario is typical in HitL robotics applications.

Our method aims to determine when the agent should request expert assistance while making the most of a limited number of such calls available during deployment. Furthermore, we will like to limit (or eliminate) the number of calls made to the expert during its training phase.

To achieve this goal, our method leverages the uncertainty of the outcomes of the agent during exploration. Specifically, we use the variance of the return from a given state as a measure of uncertainty. Intuitively, a state with high variance of the return is one from which the agent, acting alone, could achieve a range of outcomes, ranging from very poor to very effective. We posit that these are the best states in which to request assistance. In contrast, states where the variance is low are those in which the agent is certain of the outcome of the task, and there are fewer ways in which external assistance can be of help.

Most RL algorithms work by estimating the expected return from a given state; this estimate is continuously refined during training time. We modify this training procedure also to maintain and refine an estimate of $\texttt{Var}_\pi(R)$, the variance of the returns under the learned policy, as an indicator of the agent's uncertainty. At test time, our agent can use this estimate of variance to decide when to request help. We detail these processes next.

\subsection{Estimation for the variance of return}

The expected return from a given state is encapsulated by the state value functions, which are explicitly relied on by most RL algorithms. However, while most RL algorithms are not concerned with the variance of the return, we would like to compute and maintain similar estimates for it during training. By definition, the variance of the return is defined by: 
\[
\texttt{Var}(R) = \mathbb{E}[R^2] - \mathbb{E}[R]^2
\]

The expected return of taking action $a_t$ in state $s_t$ is given by the Q-function under policy $\pi$:
\begin{align}
    \mathbb{E}[R] = Q^{\pi}(s_t, a_t)
    \label{eq:qfunc}
\end{align}
Since the transition function is not available, it is usually approximated via Bellman update:

\begin{align}
    Q^{\pi}(s_t, a_t) \leftarrow (1 - \alpha) &Q^{\pi}(s_t, a_t) + \alpha(r(s_t, a_t) \nonumber\\ &+ \gamma\max_{a_{t+1}\in A}Q(a_{t+1}, s_{t+1})) \label{eq:qup}
\end{align}

The second moment of return $\mathbb{E}[R^2]$, which we call $M^{\pi}(s_t, a_t)$, can be approximated using a sampling-based method by:
\begin{align}
     M^{\pi}&(s_t, a_t) = \mathbb{E}[R^2] \nonumber \\ 
     &= \mathbb{E}[[r(s_t, a_t) + \gamma\sum_{k=t+1}^{k=N}r(s_k, a_k)]^2] \nonumber \\
     &= \mathbb{E}[r^2(s_t, a_t) + 2\gamma\sum_{k=t}^{k=N}r(s_k, a_k) + (\sum_{k=t+1}^{k=N}r(s_k, a_k))^2] \nonumber \\
     &= r^2(s_t, a_t) + 2\gamma \sum_{s_{t+1} \in S} p(s_{t+1}|s_t, a_t) Q^{\pi}(s_{t+1}, a_{t+1}) \nonumber \\
      & \hspace{1em}+ \gamma^2\sum_{s_{t+1} \in S} p(s_{t+1}|s_t, a_t) M^{\pi}(s_{t+1}, a_{t+1})
\end{align}
Since the transition function is not available for a model-free agent, we can estimate $M^{\pi}$ using a Bellman-like formula with sampled data:
\begin{align}
    \vspace{2em}
    M^{\pi}(s_t, a_t) &\leftarrow (1 - \alpha)M^{\pi}(s_t, a_t) + \alpha M'(s_t, a_t) \label{eq:mup}
    \intertext{where} \nonumber 
    M'(s_t, a_t) &= r^2(s_t, a_t) + 2\gamma Q^{\pi}(s_{t+1}, a_{t+1})  \nonumber \\
    &+ \gamma^2 M(s_{t+1}, a_{t+1})
    \label{eq:mfunc}
\end{align}
Here, $a_{t+1} = \argmax_{a_{t+1}\in A}Q(s_{t+1}, a_{t+1})$, since we assume a greedy agent that will take the actions that lead to max expected return during deployment.

Finally, combining $M$ and $Q$, we can approximate the variance of returns in a state by:
\begin{align}
        \texttt{Var}(R) = M(s_t, a_t) - Q^2(s_t, a_t)
        \label{eq:var}
\end{align}

Hence, the variance of the return given a state and action pair only depends on the $Q$ function as well as the $M$ (second function), which in turn depends on $Q$ and can be updated via a Bellman-like recursion. 

\subsection{HULA: Complete method}

\begin{algorithm}[t]
\caption{HULA Training}
\label{alg:uncertain}
\begin{algorithmic}[1]
    \WHILE {returns have not converged}
        \STATE Collect $H$ trajectories using the current control policy $\pi$ with an MDP\\
        \STATE Update estimator of the Q-function $Q_{\theta_1}$ using Eq. (\ref{eq:qup})\\
        \STATE Update estimator of the M-function $M_{\theta_2}$ using Eq.~(\ref{eq:mfunc})\\
        \STATE With updated $M$ and $Q$, compute state-dependent variance using Eq.~(\ref{eq:var}) \\
    \ENDWHILE
\end{algorithmic}
\end{algorithm}

We are now ready to integrate the variance estimation into a complete learning algorithm. We use algorithms from the Q-learning family, as these already build an explicit estimate of the $Q$ function at train time, and also use a greedy policy w.r.t. $Q$ at run-time. 

During training, in addition to the function estimator for the $Q$ function (referred to as $Q_{\theta_1}$), we build and update an estimator for $M$ (referred to as $Q_{\theta_2}$). This estimator can be of the same type as used for $Q$: for tabular Q-learning, we can use a table, while for Deep Q-Networks (DQN) \cite{Mnih2013PlayingAW} we use a deep neural network. At every iteration, we use Eqs. (\ref{eq:mup}-\ref{eq:mfunc}) to update the estimator for $M$. In the case of tabular Q-learning, Eq. (\ref{eq:mup}) can be used directly, while in the case of DQN it can be transformed in a loss function analogous to the one used for $Q$. The complete procedure is shown in Alg.~\ref{alg:uncertain}.

Finally, during deployment, the agent can use the trained estimators for the $Q$ and $M$ functions to compute the variance of the return for the current state. If this variance exceeds a threshold, the agent will request assistance and execute the action provided by the expert. Otherwise, the agent will execute the action prescribed by its own policy (in this case, a greedy selection based on the trained $Q$-function).

We note a key feature of our method: \textit{it does not require the presence of an expert during train time}. The only change at train time compared to traditional, expert-less Q-learning is the addition of the function estimator for $M$, which will be used at run-time to help provide a variance estimator. In turn, the variance estimator is used to decide when to request assistance from the expert, which is only needed at deployment.

\begin{algorithm}[t]
\caption{HULA Deployment}
\label{alg:deploy}
\begin{algorithmic}[1]
    \WHILE {not done}
        \STATE Compute variance $v_t$ in current state using Eq.~(\ref{eq:var})
        \IF{$v_t \ge \epsilon$} 
        \STATE Execute action $a_{exp}(s_t)$ provided by expert
        \ELSE
        \STATE Execute action from own policy $\pi(a_t|s_t)$
        \ENDIF
    \ENDWHILE
\end{algorithmic}
\end{algorithm}

\section{Evaluation}
\label{sec:experiments}
\begin{figure}
    \centering
    \includegraphics[width=0.98\linewidth]{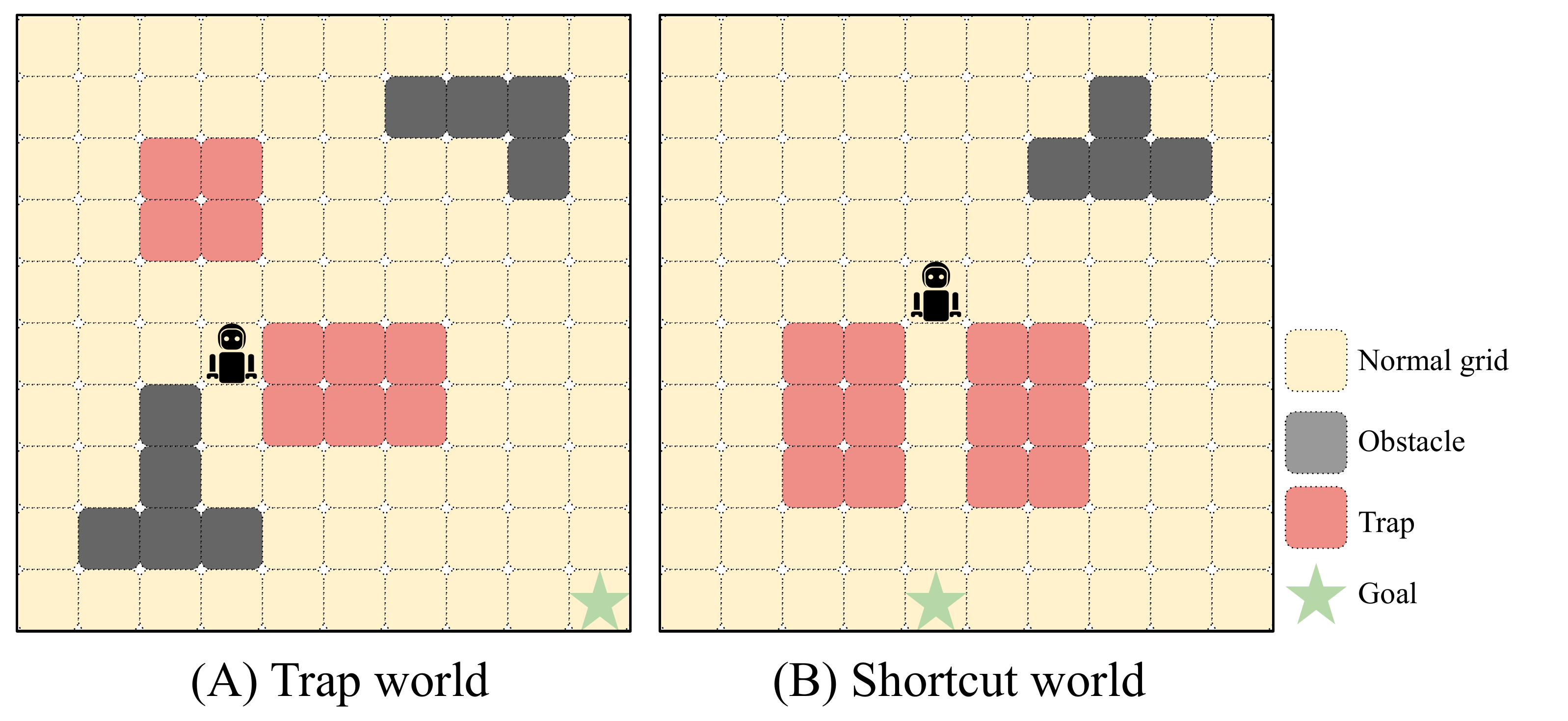}
    \caption{\textbf{Fully-observable grid worlds.}  The trap world (A) requires the agent to navigate around traps.  In the shortcut world (B), the agent can leverage the expert to use the shortcut alley to achieve higher returns.}
    \label{fig:env2-needle}
    \vspace{-1.2em}
\end{figure}
\begin{figure}
    \centering
    \includegraphics[width=0.98\linewidth]{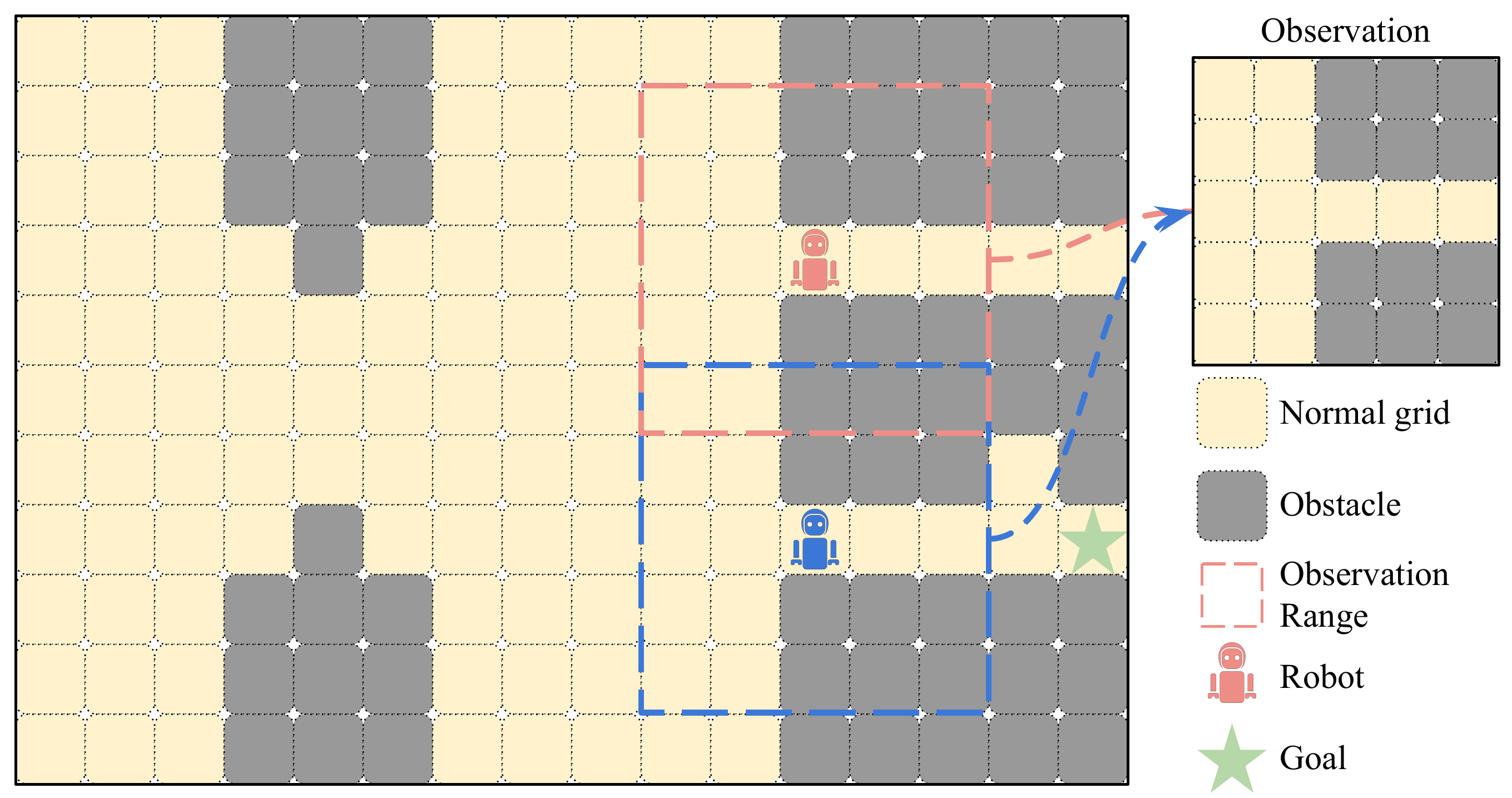}
    \caption{\textbf{Partially-observable grid world.} In this environment, an agent may not localize itself by observing a surrounding 5x5 region. For example, the blue state and the red state result in identical observations.}
    \label{fig:partial-ob}
\end{figure}

\subsection{Environments}

We evaluate HULA in a discrete navigation scenario, where the agent must reach a goal while avoiding obstacles and/or traps. Furthermore, we test our approach both on problems where the agent has exact knowledge of its current state (which is tractable via tabular Q-learning) and problems where the agent only has access to sensor data of limited range, creating ambiguity (which requires more powerful function estimators, and the use of DQN). 

\subsubsection{Fully-observable MDPs} These are discrete navigation environments where the agent is provided its exact location as observation. In addition to the goal cell, the environment also contains obstacles and traps: colliding with a trap would terminate the episode, while colliding with an obstacle results in a failure of action, and the agent would stay in its original cell. The agent also receives a small penalty for each step it takes. Available actions for an agent are moving up, down, left, and right, and the observations are the coordinates $(x, y)$ of the agent's location. We use the maps shown in Fig.~\ref{fig:env2-needle}, and train and test individually on each map.

For this simple class of problems, an unsophisticated agent can easily learn to act optimally on its own. However, we introduce uncertainty in the form of a stochastic transition function: at every step, after selecting an action, the agent moves in the desired direction with probability $\psi$, and moves in a random direction with probability $1-\psi$ (in practice, we use $\psi=0.45$). We think of this setting as a "slippery world". Thus, even with an optimal policy, the robot may not reach its goal. 

To help, we introduce an expert: when called, the expert provides the optimal action in the direction of the goal, and the action provided by the expert is not subject to transition function stochasticity. In this setting, the expert can thus be thought of as an ``action corrector'' with a better understanding of the environment dynamics, and whose actions always produce the expected results. Intuitively, we would expect an agent to call the expert in high-risk situations where it needs to avoid a wrong move, such as very close to traps. In this experiment, we test \OURS on two maps: 1. trap world that requires an agent to navigate around traps with an expert; 2. shortcut grid where the agent can leverage the expert to take the shortcut for higher returns.

\subsubsection{Partially-observable MDP's}
To test how our algorithm performs in more complex environments, we extend to a case of partially observable MDP's where the agent does not have complete information about the system. 
Instead of observing state information as the $(x, y)$ position in the grid, the observation now only includes a finite patch of grid cells around the agent's current location (we use 5x5 grids in our implementation). 

As shown in Figure \ref{fig:partial-ob}, due to partial observability, it is not always possible for the agent to uniquely identify its own location in the map from observations. In such ambiguous regions, the same action taken based on identical observations can lead to utterly different results. However, other parts of the map are uniquely identifiable based on sensor observations, and thus an autonomous agent can always select the optimal action.

In this experiment, the expert observes the full state and always provides optimal actions to the agent. Intuitively, we would expect the agent to make use of the expert when traversing ambiguous areas of the map.

\begin{figure}
    \centering
    \includegraphics[width=0.98\linewidth]{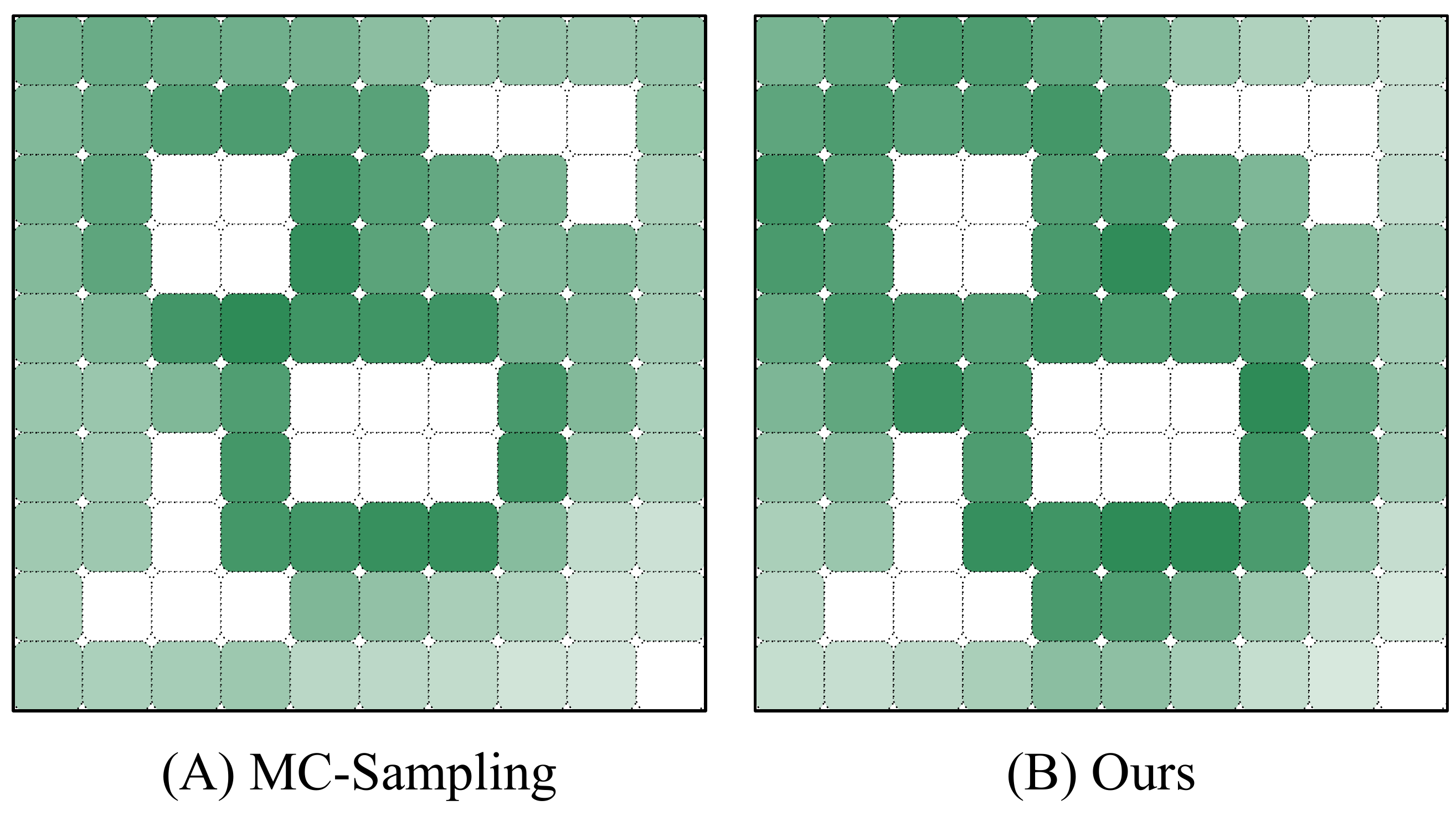}
    \caption{\textbf{A learned variance map for the trap world environment.} (A) shows the return variance computed by Monte-Carlo Sampling with the same policy. (B) shows the estimated variance by our method (darker colors represent higher variances).}
    \label{fig:env1-var}
\end{figure}
\subsection{Evaluation Approach}

The main evaluation metric for our approach focuses on its ability to make effective use of the expert: we would like to see how performance (measured as average episodic return) changes as a function of the number of expert calls made during deployment. We can measure this performance by varying the value of the variance threshold $\epsilon$ used in Alg.~\ref{alg:deploy}: for large values of $\epsilon$, the agent will never make use of the expert; conversely, if $\epsilon$ is very small, the agent will call the expert in every state. By sweeping the value of $\epsilon$ between these extremes, we obtain more or less autonomous agents, and can plot performance as a function of the number of expert calls that result in each case.

As a baseline, we use a standard RL approach that simply integrates the expert into the training procedure. We refer to this baseline as the \textit{penalty-based agent}. Specifically, the penalty-based agent treats calling expert as an extra action $a_{call}$ that can be called alongside the other actions at both training and deployment time. 

However, since the expert is optimal, the penalty-based agent will learn to always call for help. To learn a nontrivial policy that does not call the expert excessively, we employ a penalty of calling an expert in the reward function at train time: $r'(s, a) = r(s, a) + c$, where $c < 0$ is a penalty assessed only when calling an expert. By training this method with varying values of $c$, we again obtain agents that are more or less autonomous: the penalty-based agents trained with high $c$ will call the expert less often, while those trained will low $c$ will call more often. Again, we sweep the value of $c$ between these extremes, and plot performance as a function of the number of expert calls that result in each case. However, for a fair comparison against HULA, we do not assess the expert penalty at deployment time.

\section{Results}
\label{sec:results}

\begin{figure}
    \centering
    \includegraphics[width=0.9\linewidth]{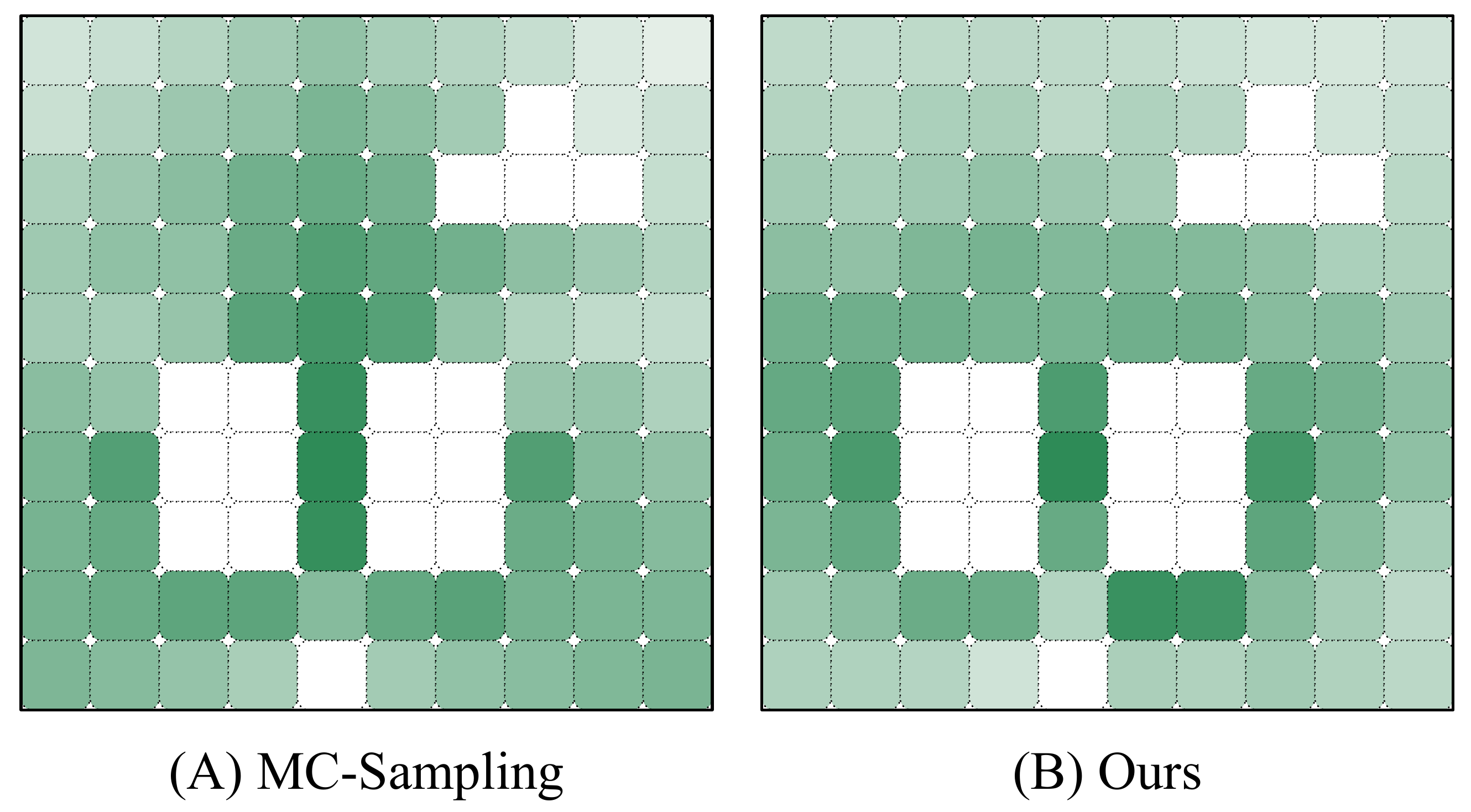}
    \caption{Comparisons between a variance map learned by \OURS and the ground truth for the shortcut world.}
    \label{fig:needle-var}
    \vspace{-1em}
\end{figure}

\begin{figure}
    \centering
    \includegraphics[width=\linewidth]{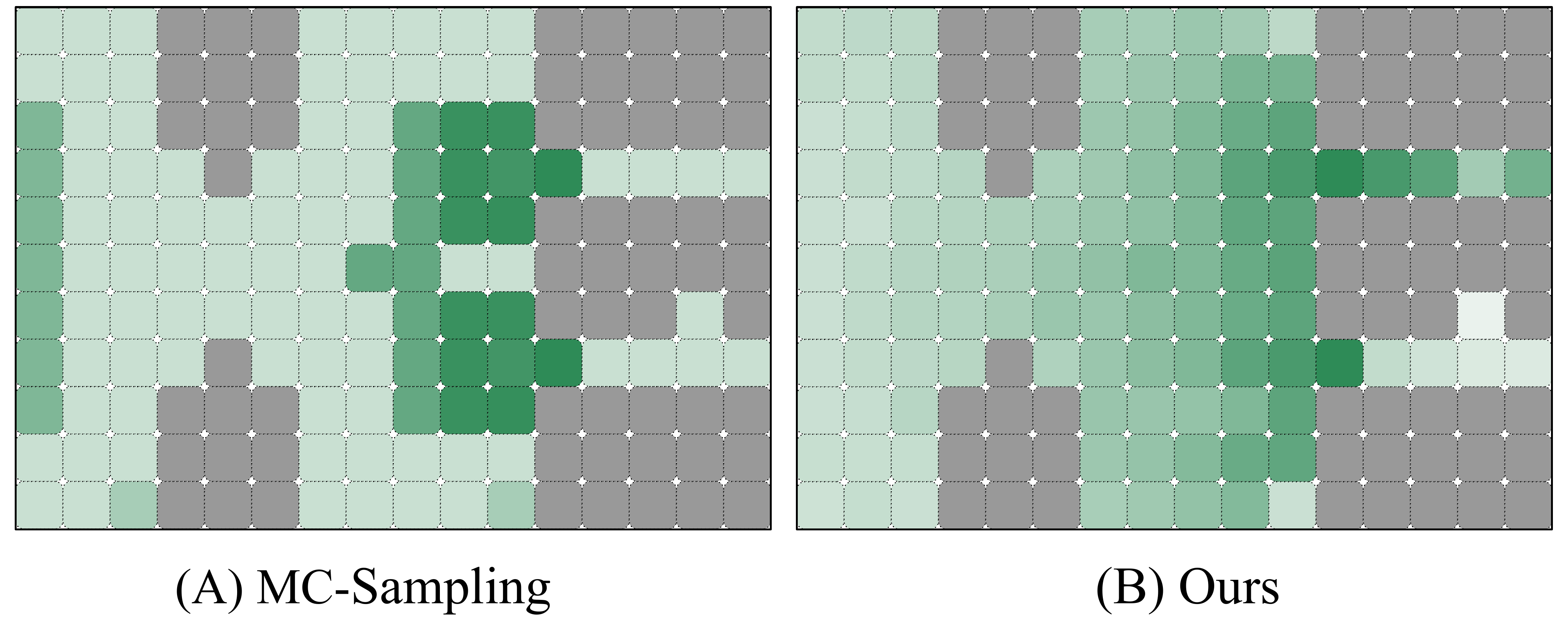}
    \caption{Comparisons between a variance map learned by \OURS and the ground truth for the partially-observable environment.}
    \label{fig:po}
        \vspace{-1em}
\end{figure}

\subsection{Uncertainty Estimation}
\begin{table}
\centering
    \caption{\textbf{Quantitative results of the expert call efficiency.} Accuracy of top-$N$ variance states estimation. PO represents the partially-observable experiments.} 
    \begin{tabular}{cccc} 
    \toprule
    Experiment & Trap world & Shortcut world & PO \\
    \midrule
    Top-5 & 0.8 & 0.2 & 0.8 \\
    Top-10 & 0.8 & 0.6 & 0.7  \\
    \bottomrule
    \end{tabular}
    \label{tab:expert-call-allocation}
\end{table}

The first question we would like to answer is whether \OURS successfully captures the uncertainty of the agent and allocate the expert calls efficiently. 
We compare the learned variances by our method with the ground truth variance of returns by Monte Carlo sampling the environment with the same policy without the assistance of an expert.

As shown in Figure \ref{fig:env1-var}, in the fully-observable experiment, \OURS successfully estimates a variance map that is similar to the ground truth. Specifically, it is uncertain about its outcome in states close to traps and confident when it is far from traps. 

In the partially observable case, \OURS identifies ambiguous states when the agent cannot localize itself in the map. In these states, the agent fails to produce an action faithfully in that the same action taken can lead to different results (e.g., hitting the boundary or reaching the goal). Our results align with the ground-truth variance map, which also has high uncertainty in a similar region.

To further investigate \OURS's efficiency in expert allocation, we compare the states with the highest $N$ variances with those in the ground truth variance map. 
These states are usually where the agent calls the expert, especially when the agent has a budget of $N$ expert calls.

Our results show that \OURS can estimate the uncertainty of an RL agent accurately and allocate its expert call based on the return variance efficiently.
As shown in Table~\ref{tab:expert-call-allocation}, in trap world, $80\%$ of the top-$5$ and top-$10$ variance states estimated by \OURS are also the top-$5$ and $10$ accordingly in the ground-truth variance. In the shortcut world, although our method only correctly estimates $1$ of $5$ highest variance states, its performance is higher and reaches $60\%$ in the top-$10$ case. This indicates that it allocates expert calls efficiently if given more budget. In the partially-observable experiment, our agent also recovers most of the high variance states in both the top-$5$ and top-$10$ evaluation.

\begin{figure}
    \centering
    \includegraphics[width=\linewidth]{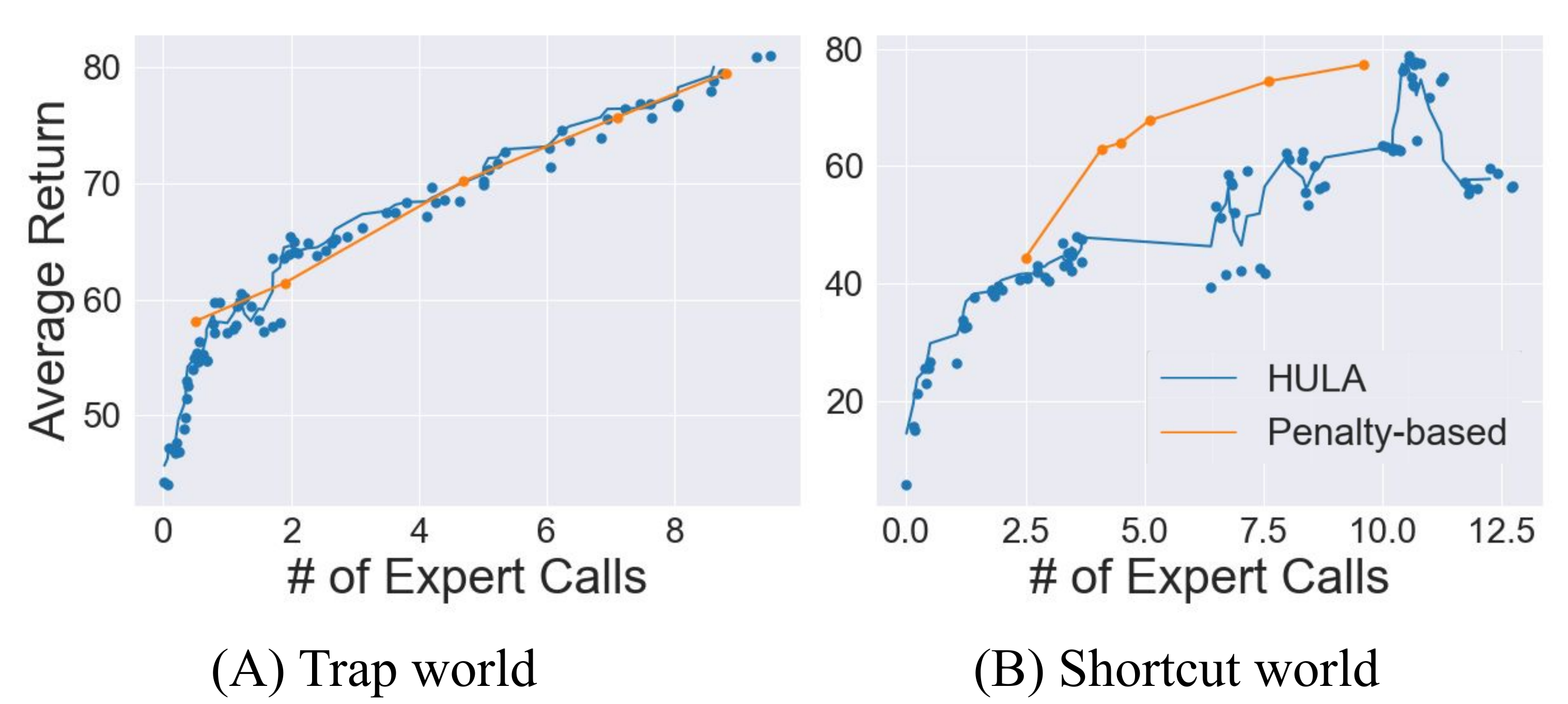}
    \caption{\# of expert calls vs Average returns for the fully-observable experiments. Curves are generated with the rolling mean method (window size of 4) for visualization purposes.}
    \label{fig:fo-performance}
\end{figure}

\subsection{Task Performance}

\begin{figure}
    \centering
    \includegraphics[width=0.9\linewidth]{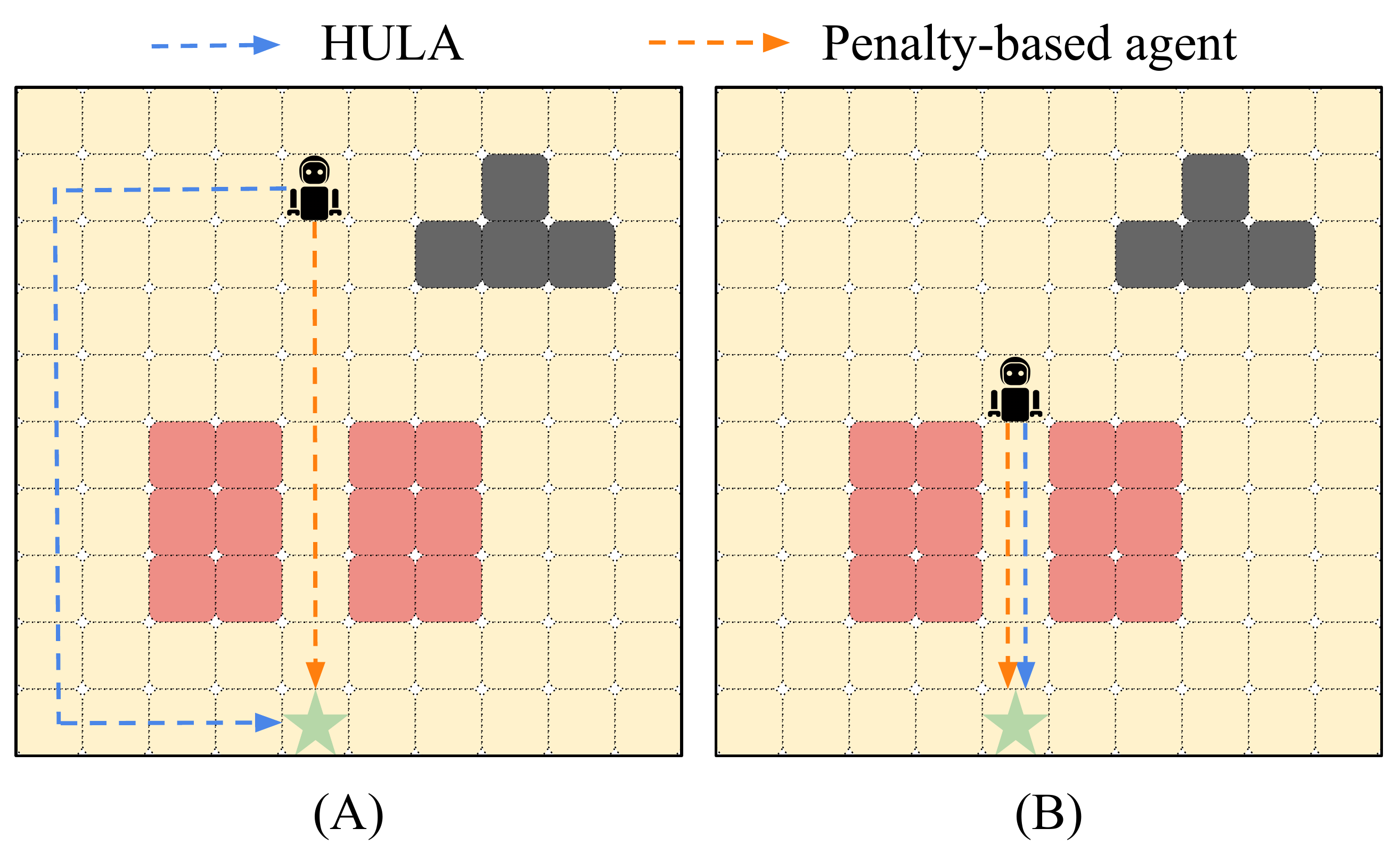}
    \caption{Example rollouts of the \OURS agent and the penalty-based agent in the shortcut world. }
    \label{fig:rollout}

\end{figure}

In all experiments, our results show that \OURS agents achieve higher performance when requesting help from experts compared to the ones that does not use an expert.
In the trap world, as shown in Fig.~\ref{fig:fo-performance}, \OURS achieves similar performance as the penalty-based agent while using the same number of expert calls. 
Interestingly, in the shortcut world, we find that the penalty-based method has a higher average return than \OURS. 
This is because our method does not evaluate the effect of calling an expert during training. As shown in Fig.~\ref{fig:rollout}, the penalty-based agent learns to move towards states that ask for help from an expert since it knows it will call an expert in a future state, whereas \OURS navigates around the uncertain states to avoid them. However, in the highest-variance states (e.g., states between traps), both agents are able to call the expert and complete the task.

In this experiment, we find that \OURS is robust to different expert call budgets compared to the penalty-based agent. 
For example, in the shortcut world, the penalty-based agent cannot learn a policy that uses 2 expert calls to complete the task even if we perform hyper-parameter sweeping with the expert penalty $c$. This can be caused by the high expert penalty that modifies the original reward structure and hinders the learning of the task.
In contrast, our method can flexibly incorporate different expert call budgets and maintain reasonable performance.

In the partially-observed grid world, the \OURS agent outperforms the penalty-based agent when the allowed expert call is limited in the range of $(2, 8]$.  When both agents are given enough expert assistance, they achieve similar performance. This indicates that \OURS efficiently utilizes the expert to localize itself and improve the task performance.

An important feature of our method is that it does not introduce extra complexity in training an RL agent.
Practically, we stop training when the $Q$ function converges. Therefore, the training efficiency is similar to its underlying RL algorithm.
In contrast, the penalty-based agent requires access to an expert during training time. Our experiments show that training each penalty-based agent for the fully-observable environment results in about $70000$ expert calls. The requirement of an expert during train time limits its application on learning policies that interact with humans.

\begin{figure}
    \centering
    \includegraphics[width=0.8\linewidth]{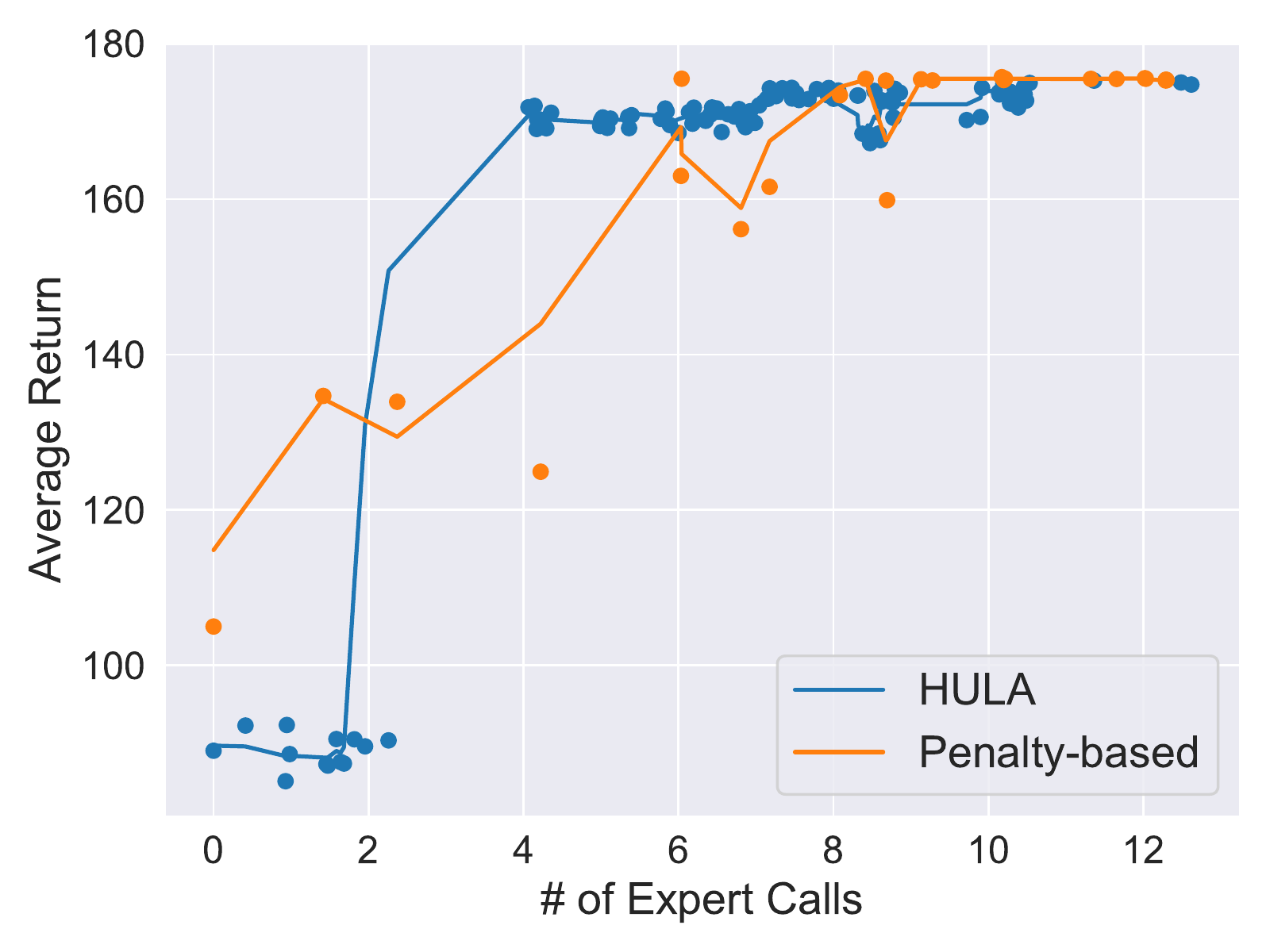}
    \caption{\# of expert calls vs Average returns for the partially-observable experiments. Curves are generated with the rolling mean method (window size of 4) for visualization purposes.}
    \label{fig:grid-results}
\end{figure}

\section{Conclusion}
\label{sec:conclusion}

We introduced \OURS, a method to learn human-in-the-loop policies by estimating an RL agent's uncertainty. 
We proposed a return variance estimation method that captures an RL agent's uncertainty. Our experimental results demonstrate that \OURS can capture an RL agent's uncertainty and use it to request assistance from experts to achieve high task performance. An important feature of our method is that it does not require the presence of an expert during training. We envision that our approach can be applied to more complex problems. For example, we plan to extend \OURS to continuous RL algorithms (e.g. DDPG \cite{Lillicrap2015ContinuousCW}) and learns HitL policies to solve continuous control problems. We also would like to explore the direction of learning uncertain agents in other domains, such as language-guided navigation.

\printbibliography

\addtolength{\textheight}{-12cm}   
\end{document}